\documentclass{article}

\usepackage[final]{neurips_2019}

\usepackage[utf8]{inputenc} 
\usepackage[T1]{fontenc}    
\usepackage{hyperref}       
\usepackage{url}            
\usepackage{booktabs}       
\usepackage{amsfonts}       
\usepackage{nicefrac}       
\usepackage{microtype}      
\usepackage{tikz}
\usetikzlibrary{arrows}
\usepackage{mathtools}
\usepackage{subfigure}
\usepackage{array}
\newcolumntype{C}[1]{>{\centering\arraybackslash}p{#1}}

\DeclareMathOperator{\Ot}{\mathcal{O}_t}

\title{DADI: Dynamic Discovery of Fair Information with\\ Adversarial Reinforcement Learning}

\author{
  Michiel A.~Bakker\\
  Massachusetts Institute of Technology\\
  MIT-IBM Watson AI Lab\\
  \texttt{bakker@mit.edu}\\
  \And 
  Duy Patrick Tu\thanks{Duy Patrick Tu did this work while visiting MIT from LMU Munich.} \\
  Massachusetts Institute of Technology\\
  Ludwig-Maximilians-Universität München\\
  \texttt{patrick2@mit.edu}\\
  \AND 
  Humberto Riverón Valdés\\
  Massachusetts Institute of Technology\\
  MIT-IBM Watson AI Lab\\
  \texttt{hriveron@mit.edu}\\
  \And
  Krishna P. Gummadi\\
  Max Planck Institute for\\ 
  Software Systems\\
  \texttt{gummadi@mpi-sws.org}\\
  \AND
  Kush R. Varshney\\
  IBM Research\\
  MIT-IBM Watson AI Lab\\
  \texttt{krvarshn@us.ibm.com}\\
  \And 
  Adrian Weller\\
  University of Cambridge\\
  Alan Turing Institute\\
  \texttt{aw665@cam.ac.uk}\\
  \And
  Alex ‘Sandy’ Pentland\\
  Massachusetts Institute of Technology\\
  MIT-IBM Watson AI Lab\\
  \texttt{pentland@mit.edu}\\
}

\begin{document}

\maketitle

\begin{abstract}
  We introduce a framework for dynamic adversarial discovery of information (DADI), motivated by a scenario where information (a feature set) is used by third parties with unknown objectives. We train a reinforcement learning agent to sequentially acquire a subset of the information while balancing accuracy and fairness of predictors downstream. Based on the set of already acquired features, the agent decides dynamically to either collect more information from the set of available features or to stop and predict using the information that is currently available. Building on previous work exploring adversarial representation learning, we attain group fairness (demographic parity) by rewarding the agent with the adversary's loss, computed over the final feature set. Importantly, however, the framework provides a more general starting point for fair or private dynamic information discovery. Finally, we demonstrate empirically, using two real-world datasets, that we can trade-off fairness and predictive performance.
\end{abstract}

\section{Introduction}
There are two parties involved in information transfer: a \emph{data owner} who has ownership over its own data or data it holds on behalf of others and a \emph{data collector} who is tasked with collecting the most informative set of data, often to maximize the performance of some predictor downstream. Intentionally or otherwise, this process of data collection and prediction can lead to biases that unfairly favor one protected subgroup over another. Numerous recent studies have shown that naively optimizing for predictive performance can lead to unfair prediction outcomes in high-stake domains such as criminal justice, credit assessment, recruiting, and healthcare \citep{kleinberg2016inherent,chalfin2016productivity,huang2007credit,obermeyer447}.

Consequently, the data owner faces a critical decision: if it cannot trust the data collector, which information should it share to ensure fair decision making? While the optimal strategy to maximize predictive performance is to naively share all the data available, the data owner has to be more careful when it wants to ensure that the predictions downstream are fair. Removing the sensitive attribute is the most obvious strategy, but is ineffective when the attribute is redundantly encoded in other features \citep{dwork2012fairness}. Another strategy is to first apply \emph{fair feature selection} in which one formulates an optimization problem to select a subset of features that maximizes accuracy, given a maximum unfairness constraint \citep{grgic2018beyond}. This strategy, though effective, is inefficient as it removes each feature simultaneously for all individuals, ignoring any differences in the underlying conditional dependencies. For example, for individuals that live in Chicago, the most racially segregated city in America, zipcode will be highly correlated with race and using this feature can thus lead to racially biased predictions \citep{logan2014diversity}. In contrast, if an individual lives in Irvine, California, America's most racially integrated city, zipcode alone will not reveal an individual's race. Removing zipcode for all individuals is therefore an effective but inefficient strategy to ensure fairness.

Motivated by this problem, we propose the DADI (Dynamic Adversarial Discovery of Information) framework as a general sequential information acquisition framework for any task. Our contributions are as follows: to the best of our knowledge, we introduce the first framework for dynamic adversarial discovery of information which we utilize to acquire feature sets that ensure fair decision making. In this framework, we formulate the feature acquisition task as a minimax optimization problem in which a reinforcement learning (RL) agent simultaneously minimizes the classification loss while maximizing the loss of an adversary. We actualize this with a joint framework that simultaneously trains a classifier, an adversary, and an RL agent using deep Q-learning. Building on work on adversarial representation learning, we investigate the effects of two different adversarial reward functions to achieve \emph{demographic parity} \citep{edwards2015censoring,madras2018learning}. Finally, we demonstrate the effectiveness of our framework with two real-world public datasets.

\section{Related Work}
\paragraph{Fairness} Recent years have seen an explosion in academic work that seeks to define and obtain fairness in automated decision making systems. At a high level, this literature has focused on two families of definitions: \emph{statistical} notions of fairness and \emph{individual} notions of fairness \citep{dwork2012fairness,verma2018fairness}. Most of the literature, including this work, focuses on statistical or group definitions of fairness, in which we require parity of some statistical measure to hold across a small number of protected subgroups. In contrast, individual fairness definitions have no notion of protected subgroups, but instead formulate constraints that bind on pairs of individuals \citep{dwork2012fairness,joseph2016fairness}. Both families of definitions have strengths and weaknesses; statistical notions are easy to verify but do not provide any guarantees to individuals, while individual notions do give individual guarantees but are difficult to implement in practice and are ambiguous with respect to the agreed-upon distance function. 

In this work, we focus on \textit{demographic parity}, requiring parity of the positive classification rate across groups, i.e. $P(\hat{y}=1\mid b=0)=P(\hat{y}=1\mid b=1)$, where $\hat{y} \in \{ 0,1\}$ is the binary prediction of a model that classifies feature set $\mathbf{x}$ and $b \in \{ 0,1\}$ is the sensitive attribute. The usefulness of demographic parity is limited when the base rate differs across groups, i.e. $P({y}=1\mid b=0)\ne({y}=1\mid b=1)$ where ${y} \in \{ 0,1\}$ is the ground truth label. In that case, the metric can be generalized by conditioning on the ground truth label, yielding equal false negative rates (\emph{equal opportunity}) or equal false negative and false positive rates (\emph{equal odds}) as measures of fairness \citep{hardt2016equality}. We demonstrate the effectiveness of our framework using demographic parity, but note that alternative adversarial objectives have been introduced that can be combined with our framework to achieve equal opportunity or equal odds \citep{madras2018learning}.

\paragraph{Adversarial training} Adversarial training for deep generative models was introduced in \citet{goodfellow2014generative}, framing the learning as a two-player game between a generator and a discriminator. The generator aims to fool the discriminator by generating fake data that resembles data from a dataset $X$ while the discriminator is trained to distinguish between `real' data from and `fake` data generated by the generator. Learning proceeds using a minimax optimization where the generator and discriminator are optimized jointly. At each iteration, the discriminator improves its ability to discriminate between real and fake which, in turn, forces the generator to generate fake data that better resembles the real data.

Adversarial training was first applied in the context of fairness by \citet{edwards2015censoring}, proposing adversarial training to ensure that multiple distinct data distributions from different demographic subgroups are modeled as a single representation. The discriminator aims to distinguish between subgroups while an encoder aims to map each data distribution to a single representation to fool the discriminator. Subsequently, these representations can be safely shared with a data collector while ensuring demographic parity for predictions downstream. \citet{beutel2017data} further explores this approach in the context of demographically imbalanced data. Finally, \citet{zhang2018mitigating} and \citet{madras2018learning} extend this body of work by connecting multiple statistical notions of fairness to different adversarial objectives. Whereas the method presented in \citet{zhang2018mitigating} predicts the sensitive attribute from the prediction of the classifier, our work is closer to the method in \citet{madras2018learning}, working directly with the learned representation. This allows for transferable representations that ensure fair outcomes for other third-party classifiers downstream. 

Although this work is similar in spirit to adversarial representation learning, we aim to dynamically collect a fair subset of features instead of learning to map the full feature set to a fair representation. The ability to collect raw features instead of mapping to a representation is crucial for integration with current information systems where the collected information is used or audited by both human and machine decision makers downstream. If we consider our the example of credit assessment, a bank not only wants to collect a low-level abstract representation for the purpose of the initial creditworthiness prediction but also wants to explain the credit decisions to an applicant and store the applicant's information in a database to provide other services downstream or allow for audits.

\paragraph{Active feature-value acquisition} Different from \emph{active learning}, active feature-value acquisition (AFA) is concerned with feature-wise active learning for each instance. AFA is of great need in cost-sensitive applications where the data collector needs to balance an available information budget with predictive accuracy.  A traditional AFA system consists of three components: 1) a classifier that can handle partially observed feature sets, 2) a strategy for determining which feature to select next based on the features that are already collected, and 3) a stopping criterion for determining when to stop acquiring more features and make a final prediction.

First, there are different ways a classifier can handle a partial features set. Generative models handle missing features naturally by first integrating out the variables while in discriminate models feature imputation or expectation-maximization can be used to first replace the missing values with estimates.  In this work, we use a set encoder based on \citet{vinyals2015order} to encode arbitrary subsets of features. Second, to determine which feature to select next, we need a method that estimates the value of each of the unselected features based on the features that we have already collected. A recent approach, Efficient Dynamic Discovery of High-Value Information (EDDI), uses a partial variational autoencoder to represent the set of already acquired features. It then computes the mutual information between the current representation and each of the available features to select the feature that minimizes this information \citep{ma2018eddi}. Finally, a stopping criterion is not specified in EDDI and most other AFA methods. However, some prior work assumes a fixed feature budget per individual after which the process terminates \citep{krishnapuram2011cost}. The \emph{active fairness} framework presented in \citet{noriega2018active} extends this to group-specific budgets that are found to attain equal opportunity (equal false-positive or false-negative rates).

To effectively trade-off fairness and accuracy, we need a unified framework that jointly optimizes both the acquisition strategy and the stopping criterion. We adopt the framework from \citet{shim2018joint} and model the feature acquisition process as a Markov decision process (MDP) where the action space consists of the set of unselected set of features and an additional STOP action which, upon selection, terminates the acquisition process. To ensure fairness, we formulate a reward function that balances low classification loss with a high adversarial loss.  
\newcommand{\arrowWidth}{1.5pt}
\newcommand{\rectWidth}{0.5cm}
\newcommand{\circleSize}{0.9cm}
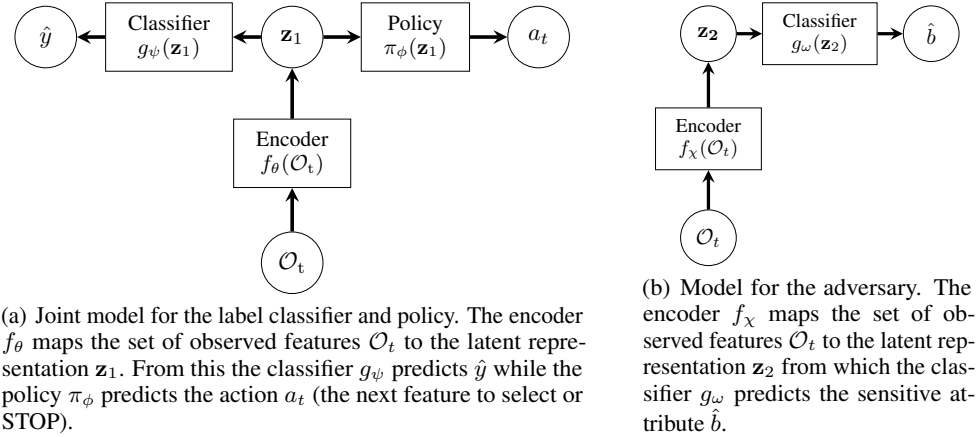
\begin{figure}
\begin{tabular}{C{.58\textwidth}C{.34\textwidth}}
    \subfigure[Joint model for the label classifier and policy. The encoder $f_\theta$ maps the set of observed features $\mathcal{O}_t$ to the latent representation $\mathbf{z}_1$. From this the classifier $g_\psi$ predicts $\hat{y}$ while the policy $\pi_\phi$ predicts the action $a_t$ (the next feature to select or STOP).]{
        \resizebox{0.54\textwidth}{!}{
            \begin{tikzpicture}
                \path  (6.6,3.3) node[circle,draw,minimum size=\circleSize,align=center](a) {$a_t$}
                (3,3.3) node[circle,draw,minimum size=\circleSize,align=center](z) {$\mathbf{z}_{1}$}
                (-0.6,3.3) node[circle,draw,minimum size=\circleSize](y) {$\hat{y}$}
                (3,0) node[circle,draw,minimum size=\circleSize,align=center](O) {$\Ot$}
                (3,1.6) node[rectangle,draw,minimum width=\rectWidth,minimum height=1cm](encoder) {\footnotesize \begin{tabular}{c} Encoder \\ $f_\theta(\Ot)$ \end{tabular}}
                (1.2,3.3) node[rectangle,draw,minimum width=\rectWidth,minimum height=1cm](classifier) {\footnotesize \begin{tabular}{c} Classifier \\ $g_\psi(\mathbf{z}_{1})$ \end{tabular}}
                (4.8,3.3) node[rectangle,draw,minimum width=\rectWidth,minimum height=1cm](policy) {\footnotesize \begin{tabular}{c} Policy \\ $\pi_\phi(\mathbf{z}_{1})$ \end{tabular}};
                \draw[->,>=stealth, line width=\arrowWidth] (O) -- (encoder);
                \draw[->,>=stealth, line width=\arrowWidth] (encoder) -- (z);
                \draw[->,>=stealth, line width=\arrowWidth] (z) -- (classifier);
                \draw[->,>=stealth, line width=\arrowWidth] (classifier) -- (y);
                \draw[->,>=stealth, line width=\arrowWidth] (z) -- (policy);
                \draw[->,>=stealth, line width=\arrowWidth] (policy) -- (a);
            \end{tikzpicture}
        }
    } &
    \subfigure[Model for the adversary. The encoder $f_\chi$ maps the set of observed features $\mathcal{O}_t$ to the latent representation $\mathbf{z}_2$ from which the classifier $g_\omega$ predicts the sensitive attribute $\hat{b}$.]{
        \resizebox{0.30\textwidth}{!}{
            \begin{tikzpicture}
            \path (2,0) node[circle,draw,minimum size=\circleSize](O) {$\mathcal{O}_t$}
            (2,3.3) node[circle,draw,minimum size=\circleSize](z) {$\mathbf{z_2}$}
            (5.6,3.3) node[circle,draw,minimum size=\circleSize](y) {$\hat{b}$}
            (2,1.6) node[rectangle,draw,minimum width=\rectWidth,minimum height=1cm](encoder) {\footnotesize \begin{tabular}{c} Encoder \\ $f_\chi(\mathcal{O}_t)$ \end{tabular}}
            (3.8,3.3) node[rectangle,draw,minimum width=\rectWidth,minimum height=1cm,align=center](classifier) {\footnotesize \begin{tabular}{c} Classifier \\ $g_\omega(\mathbf{z}_{2})$ \end{tabular}};
            \draw[->,>=stealth, line width=\arrowWidth] (O) -- (encoder);
            \draw[->,>=stealth, line width=\arrowWidth] (encoder) -- (z);
            \draw[->,>=stealth, line width=\arrowWidth] (z) -- (classifier);
            \draw[->,>=stealth, line width=\arrowWidth] (classifier) -- (y);
            \end{tikzpicture}
        }
    }
\end{tabular}
\caption{Joint framework for dynamic adversarial discovery of information (DADI)}\label{fig:model}
\end{figure}
\section{Adversarial Discovery of Fair Information}
\paragraph{Problem setup} The setup of our framework most follows the joint active feature acquisition and classification framework in \citep{shim2018joint}; however, we extend their framework for use with an adversary. Let $(\mathbf{x}^{(i)},y^{(i)},b^{(i)})\sim P$ be individual $i$ in $P$ represented by a $d$-dimensional feature vector $\mathbf{x}^{(i)} \subseteq \mathbb{R}^d$, a binary label $y^{(i)} \in \{0,1\}$, and a binary sensitive attribute $b^{(i)} \in \{0,1\}$. We acquire the features in sequential order starting with an empty set $\mathcal{O}_0:=\emptyset$ at time $t=0$. At every later timestep $t$, we choose a subset of features from the unselected set of features, $\mathbf{S}_t^{(i)} \subseteq \{1,\dots,d\} \setminus \mathcal{O}_{t-1}^{(i)}$. After each new acquisition step, the classifier will have access to feature values in $\mathcal{O}_t^{(i)} := \mathcal{S}_{t}^{(i)} \cup \mathcal{O}_{t-1}^{(i)}$. We keep acquiring features up to time $T^{(i)}$ when we meet a stopping criterion. At that point, we will classify $\mathbf{x}^{(i)}$ using only the set of features in $\mathcal{O}_{T^{(i)}}^{(i)}$. Note that the specific set of selected features $\mathcal{O}_{T^{(i)}}^{(i)}$ will generally be different for each individual $i$. To learn the model that minimizes classification loss while maximizing the loss of the adversary we formulate the following optimization problem. 
\begin{align} \label{eq:lossfunc}
    \underset{\psi, \theta,\omega,\chi}{\text{minimize}} \frac{1}{|P|} \sum_{i\in P} (1-\gamma)\mathcal{L}_C\left(g_\psi(f_{\theta}(\mathcal{O}_T^{(i)}),y^{(i)}\right) - \gamma \mathcal{L}_A\left(g_\omega(f_{\chi}(\mathcal{O}_T^{(i)}),b^{(i)}\right)
\end{align}
Where $\mathcal{L}_C$ and $\mathcal{L}_A$ are the suitable losses for the label classifier and the adversary. The encoder $f_{\theta}$ feeds into a classifier $g_\psi$ for the label prediction $\hat{y}$ while $f_{\chi}$ and $g_\omega$ are the encoder and classifier for the sensitive attribute prediction $\hat{b}$. Hyperparameter $\gamma$ specifies the desired balance between classification performance and fairness. When clear from context, we drop the superscript $(i)$.

\paragraph{Markov decision process} We define a Markov decision process (MDP) to find the set of features $\mathcal{O}_T^{(i)}$ that minimizes the objective in Eq. \ref{eq:lossfunc}. For each episode, the state at time $t$ is represented by the set of selected features $\{x_j\}_{j \in \mathcal{O}_t}$. The size of the state space is $2^d$, the powerset of the feature set. At each timestep $t$, the action space consists of the set of unselected features $\{1,\dots,d\} \setminus \mathcal{O}_{t-1}$ and an additional STOP action which, upon selection, stops the acquisition process after which the rewards are computed. The agent's reward function, computed at end of the episode for individual $i$, corresponds to
\begin{align}
    r(\mathcal{O}_T^{(i)})= {-(1-\gamma)\mathcal{L}_C\big(g_\psi(f_{\theta}(\mathcal{O}_T^{(i)}),y^{(i)})\big)}_{}+{\gamma \mathcal{L}_A\big(g_\omega(f_{\chi}(\mathcal{O}_T^{(i)}),b^{(i)})\big)}_{} \label{eq:rewards}
 \end{align}
 where the first reward encourages accurate classification and the second reward encourages low mutual information between the feature set and the sensitive attribute. Now, if we now consider a policy $\pi_\phi^*$, parametrized by $\phi$, that is optimal for this MDP, then $\pi_\phi^*$ is also the optimal solution to the objective in Eq. \ref{eq:lossfunc}. We can proof this by maximizing the aggregated reward in Eq. \ref{eq:rewards} over the population $P$
\begin{align}
   &\underset{\phi}{\arg \max}\frac{1}{|P|}\sum_{i \in P} -(1-\gamma)\mathcal{L}_C\big(g_\psi(f_{\theta}(\mathcal{O}_T^{(i)}),y^{(i)})\big)+\gamma \mathcal{L}_A\big(g_\omega(f_{\chi}(\mathcal{O}_T^{(i)}),b^{(i)})\big)\\
   =&\underset{\phi}{\arg \min}\frac{1}{|P|}\sum_{i \in P} (1-\gamma)\mathcal{L}_C\big(g_\psi(f_{\theta}(\mathcal{O}_T^{(i)}),y^{(i)})\big)-\gamma \mathcal{L}_A\big(g_\omega(f_{\chi}(\mathcal{O}_T^{(i)}),b^{(i)})\big)
\end{align}
which is equivalent to the minimization objective in Eq. \ref{eq:lossfunc}.

\paragraph{Generalized framework} The generalized framework in Fig. \ref{fig:model} consists of two parts: the first part in Fig. \ref{fig:model}(a) seeks to learn a representation of the set of observed features $\mathbf{z_1}=f_\theta(\mathcal{O}_t)$ capable of classifying the label $\hat{y}=g_\psi(\mathbf{z_1})$ and estimating the optimal next action $a_t=\pi_\phi(\mathbf{z_1})$. The model has two heads that share the same encoder which leads to improved performance over a model with two separate encoders \citep{shim2018joint}. In parallel, the second network in Fig. \ref{fig:model}(b) seeks to learn a related but separate representation $\mathbf{z_2}=f_\chi(\mathcal{O}_t)$, which is fed to a classifier $g_\omega$ that predicts the sensitive attribute $\hat{b}$. Crucially and different from prior work on adversarial representation learning, the second adversarial classification task cannot have a shared encoder with the first two tasks as this could encourage the encoder to mask the unfairness of features directly which, in turn, would not lead to selecting a set of fair features that generalize to any downstream task. While in adversarial representation learning the adversarial loss is backpropagated directly through a gradient reversal layer to update the encoder \citep{goodfellow2014generative,edwards2015censoring}, our agent learns to fool the adversary by selecting the set of features that maximize the adversarial classification loss.

We realize $f_\theta$, $g_\psi$, $f_\chi$, $g_\omega$ and $\pi_\phi$ as neural networks parametrized by $\theta$, $\psi$, $\chi$, $\omega$, and $\phi$, which are optimized using alternating gradient descent steps. To facilitate encoding of partially observed feature sets, we adopt a feature-level set encoder \citep{shim2018joint}. Each observed feature $x_i$ is first mapped to a memory vector $\mathbf{m}_i$ after which an LSTM processes the set of memory vectors repeatedly while an attention layer improves the set embedding. The attention step ensures the input is order-invariant. The final set embedding $\mathbf{z}_1$ is fed to both the classifier and the policy network. A second independent set embedding $\mathbf{z}_2$ is fed to the adversary. We refer to App. \ref{sec:setencoder} for details on the set encoding process and to App. \ref{sec:arch} for implementation details.

\paragraph{Adversarial reward function} We compare two different loss functions to compute the rewards for the adversary. First, earlier work on adversarial fair representation learning for demographic parity has shown that using binary cross-entropy (CE) loss for both the classifier and the adversary encourages fair and high-value representations \citep{edwards2015censoring,beutel2017data}
\begin{align}
     \mathcal{L}_{A}^{CE}(\mathcal{O}_T) = -\left(b \log (g_\omega(f_\chi(\mathcal{O}_T))) + (1 - b) \log(1-g_\omega(f_\chi(\mathcal{O}_T)))\right)
\end{align}
where the adversary only has access to the final feature set $\mathcal{O}_T$ obtained after stopping. Though effective, $\mathcal{L}_{A}^{CE}$ fails to account for demographically unbalanced training data. To address this problem, \citep{madras2018learning} introduces group-normalized $L_1$ (GN$L_1$) loss as a more natural relaxation of demographic parity, which we adopt to compute the rewards from the adversary
\begin{align}
     \mathcal{L}_{A}^{GNL_1}(\mathcal{O}_T) = \frac{|P|}{2|P_b|} | g_\omega(f_\chi(\mathcal{O}_T)) - b |
\end{align}
where $P_0$ and $P_1$ are the protected subgroups with respectively attributes $b=0$ and $b=1$. As neural networks have difficulty learning with $L_1$ loss \citep{janocha2017loss}, we continue to use cross-entropy loss to train the adversary but use $\mathcal{L}_{A}^{GNL_1}$ to compute the final rewards for the agent. We refer to \citet{madras2018learning} for the theoretical properties of both loss functions.

\section{Experiments}
DADI seeks to select the subset of features that can be used by third parties with the assurance that their trained classifiers are both fair and accurate. As exact demographic parity is hard to enforce in practice, we use demographic disparity $\left|P(\hat{y}=1 \mid b=0) - P(\hat{y}=1 \mid b=1)\right|$ as measure for the degree of unfairness. The performance of the classifier is measured using the Area Under the Receiver Operating Characteristics curve (AUC) to account for the imbalanced label distributions.

\paragraph{Datasets} We evaluate DADI empirically on the UCI Adult and Mexican Poverty datasets. We use one-hot encoding for categorical features and standardize numerical features. For the mapping to actions, we combine multiple one-hot encoded binary features that stem from the same categorical feature into a single action (e.g. the binary features \textit{marital=divorced, marital=married} and \textit{marital=single} correspond to a single action that acquires these features simultaneously). We use 8-fold cross validation with a random $87.5\%/12.5\%$ train-val/test split. We further split the train-val set into training and validation data using a second random $80\%/20\%$ split. 

 The Adult Dataset from UCI Machine Learning Repository \citep{lichman2013uci} comprises 14 demographic and occupational attributes, which translates after preprocessing into 98 continuous and binary features and 14 actions for 48,842 individuals, with the goal of classifying whether a person’s income is above \$50,000 (25\% are above). Rows with missing values are omitted resulting in a dataset with 45,222 samples. In line with previous work, we use gender as the sensitive attribute, listed as male or female.  
 
 The Mexican Poverty dataset is extracted from the Mexican household survey 2016, which contains ground-truth household poverty levels and 99 attributes, related to household information such as the number of rooms or the type of heating system \citep{ibarraran2017conditional}. The processed dataset is obtained from \citet{noriega2018active} and comprises a sample of 70,305 households in Mexico, with 183 continuous and one-hot encoded binary features and 99 actions. Classification is binary according to the country’s official poverty line, with 36\% of the households having the label poor. The considered sensitive attribute describes whether the head of the household is a senior citizen or not.
 
\begin{figure*}[t]
\centering
\subfigure[Adult Income: Disparity-$(1-\gamma)$]{
    \includegraphics[trim=-.5cm 0 0.2cm 0,clip, height=3.8cm]{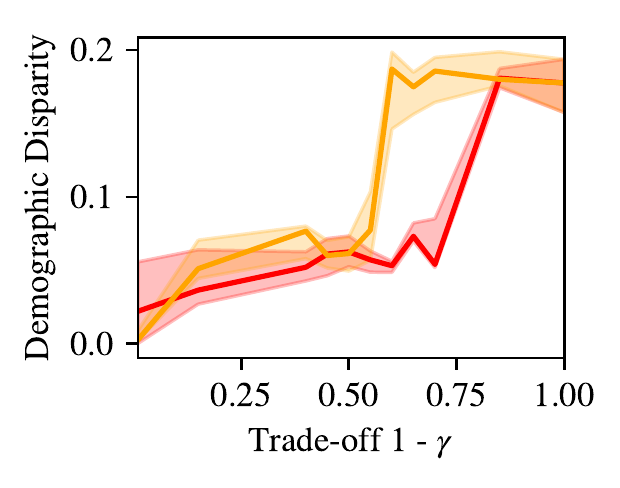}
    \label{fig:subfig1a}
}
\subfigure[Adult Income: AUC-$(1-\gamma)$]{
    \includegraphics[trim=0.2cm 0 0.2cm 0,clip, height=3.8cm]{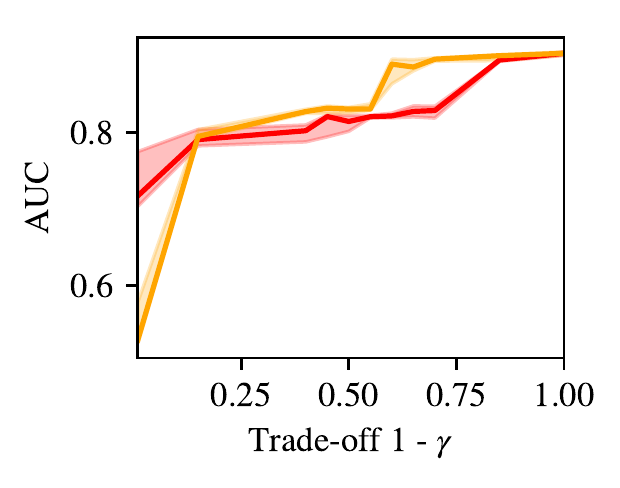}
    \label{fig:subfig2a}
}
\subfigure[Adult Income: AUC-disparity]{
    \includegraphics[trim=0.6cm 0 0.2cm 0,clip, height=3.8cm]{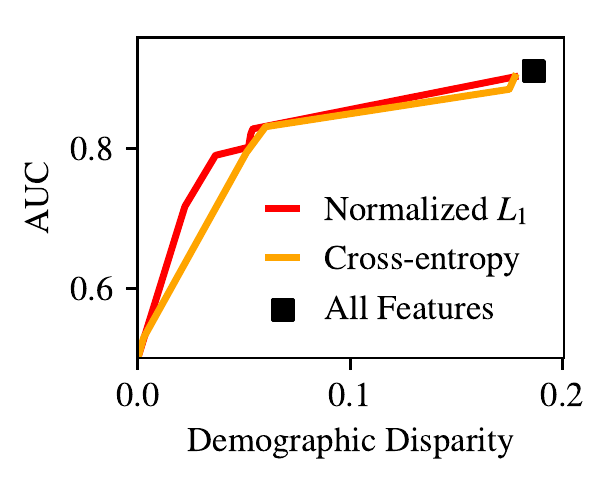} 
    \label{fig:subfig3a}
}\\
\subfigure[Mex. Poverty: Disparity-$(1-\gamma)$]{
    \includegraphics[trim=-.5cm 0 0.2cm 0,clip, height=3.8cm]{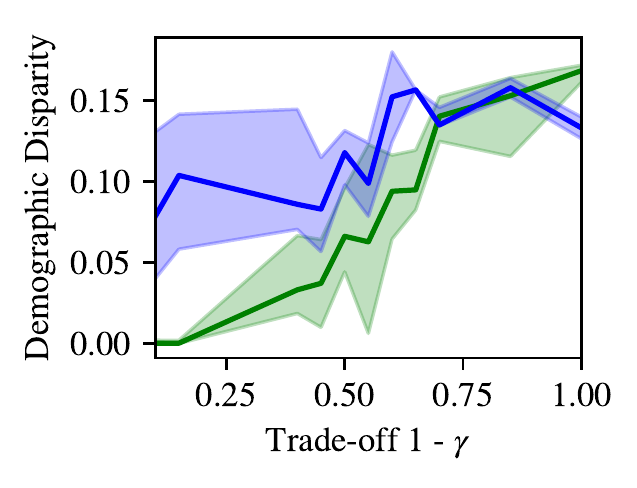}
    \label{fig:subfig1m}
}
\subfigure[Mex. Poverty: AUC-$(1-\gamma)$]{
    \includegraphics[trim=0.2cm 0 0.2cm 0,clip, height=3.8cm]{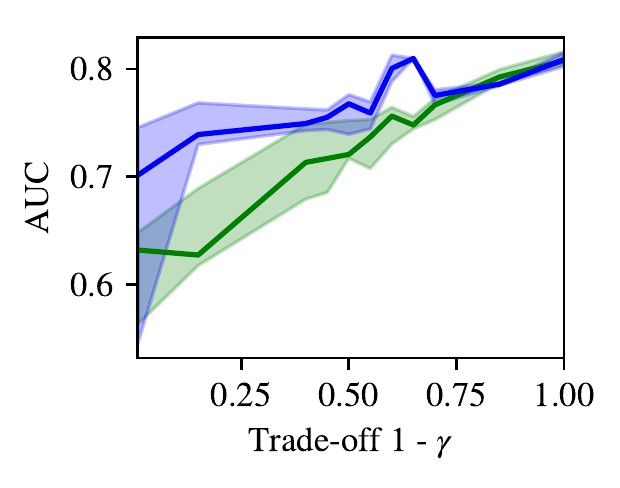}
    \label{fig:subfig2m}
}
\subfigure[Mex. Poverty: AUC-disparity]{
    \includegraphics[trim=0.6cm 0 0.2cm 0,clip, height=3.8cm]{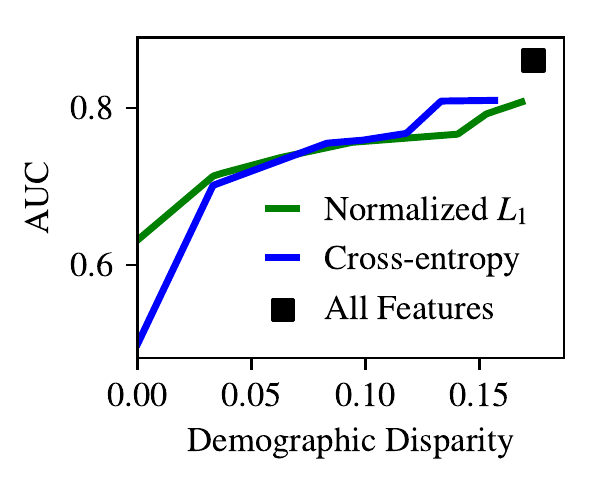} 
    \label{fig:subfig3m}
}
\caption{DADI for mitigating demographic disparity across subgroups in the Adult and Mexico datasets. Subfigures \subref{fig:subfig1a},\subref{fig:subfig2a}, \subref{fig:subfig1m} and \subref{fig:subfig2m} show respectively the AUC and disparity for a range of trade-off parameters $1-\gamma$. The lines are plotted using the median with first/third quantile as confidence area computed using 8-fold CV. Subfigures \subref{fig:subfig3a} and \subref{fig:subfig3m} show the Pareto front along the AUC-disparity trade-off. The black square represents the baseline unfair classifier for which we use the pretrained classifier together with the full feature set. The median AUC and disparity are again computed across the 8 folds.}
\label{fig:results}
\end{figure*}

\paragraph{Results}
Fig. \ref{fig:results} shows the results for both datasets. First, Figs.  \ref{fig:subfig1a},\ref{fig:subfig2a},\ref{fig:subfig1m}, and \ref{fig:subfig2m} show that increasing $1-\gamma$, i.e., decreasing the relative weight of the adversarial reward $\gamma$, leads to an increase in both performance and disparity for both choices of adversarial reward functions. Naturally, as the adversarial reward becomes less important, the agent will have a stronger incentive to maximize the accuracy which, in turn, leads to the collection of more features and thus higher AUC at the cost of a higher disparity.

Importantly, however, we observe that while the AUC increases drastically from the start, demographic parity only increases drastically for larger values of $1-\gamma$, allowing for agents that achieve good predictive performance with minimal disparity loss. This conclusion is supported by Figs. \ref{fig:subfig3a} and \ref{fig:subfig3m} where we visualize the Pareto front along the AUC-disparity trade-off. These results are encouraging as we show that a data collector can still maintain good performance while only having access to a unique fair subset of features for each data owner. Finally, we observe that the group-normalized $L_1$ reward generally results in a better trade-off, especially in the most important fairness range for small values of the disparity.

\section{Conclusion and Future Work}
A number of recent works have focused on adversarially learning fair representations. However, the methods underlying these works, are ineffective when the data owner is required to share raw features, a key aspect in many use cases where features are collected for both human and machine decision making. To tackle this problem, we propose DADI, to our knowledge the first framework for dynamic adversarial discovery of fair information. We frame the data owner's choice as a reinforcement learning problem where an agent selects a subset of features while an adversary critiques potentially unfair feature sets. Experimentally, we demonstrate how our framework guides information discovery for ensuring demographic parity and how it allows the data owner to efficiently trade-off fairness and accuracy. 

Importantly, however, our framework is more generally applicable in settings where a data owner may wish to guard itself against a naive or malicious data collector by sharing only a subset of features. First, by changing the adversarial objective function, the framework in \citep{madras2018learning} demonstrates that one can achieve other notions of fairness such as equal opportunity and equal odds. Second, several recent works have formulated adversarial objectives to attain (differentially) private data representations. These objectives could be adopted using DADI to automate dynamic discovery of private information \citep{yang2018learning,phan2019preserving} which could be further extended by encoding features in different levels of precision (such as age by year or age by decade), allowing the agent to select the level of precision that maximizes accuracy while minimizing privacy risk. Finally, adding monetary acquisition costs of features as a penalty at each collection step would allow our agent to holistically trade-off accuracy, information costs, and fairness or privacy \citep{shim2018joint}. 

\section*{Acknowledgements}
The authors want to thank IBM for access to their compute resources and Prasanna Sattigeri for helpful discussions. AW acknowledges support from the David MacKay Newton research fellowship at Darwin College, The Alan Turing Institute under EPSRC grant EP/N510129/1 \& TU/B/000074, and the Leverhulme Trust via the CFI.

\bibliography{sample-base.bib}

\begin{thebibliography}{26}
\providecommand{\natexlab}[1]{#1}
\providecommand{\url}[1]{\texttt{#1}}
\expandafter\ifx\csname urlstyle\endcsname\relax
  \providecommand{\doi}[1]{doi: #1}\else
  \providecommand{\doi}{doi: \begingroup \urlstyle{rm}\Url}\fi

\bibitem[Beutel et~al.(2017)Beutel, Chen, Zhao, and Chi]{beutel2017data}
Alex Beutel, Jilin Chen, Zhe Zhao, and Ed~H Chi.
\newblock Data decisions and theoretical implications when adversarially
  learning fair representations.
\newblock \emph{arXiv preprint arXiv:1707.00075}, 2017.

\bibitem[Chalfin et~al.(2016)Chalfin, Danieli, Hillis, Jelveh, Luca, Ludwig,
  and Mullainathan]{chalfin2016productivity}
Aaron Chalfin, Oren Danieli, Andrew Hillis, Zubin Jelveh, Michael Luca, Jens
  Ludwig, and Sendhil Mullainathan.
\newblock Productivity and selection of human capital with machine learning.
\newblock \emph{American Economic Review}, 106\penalty0 (5), 2016.

\bibitem[Dwork et~al.(2012)Dwork, Hardt, Pitassi, Reingold, and
  Zemel]{dwork2012fairness}
Cynthia Dwork, Moritz Hardt, Toniann Pitassi, Omer Reingold, and Richard Zemel.
\newblock Fairness through awareness.
\newblock In \emph{Proceedings of the 3rd innovations in theoretical computer
  science conference}, pages 214--226. ACM, 2012.

\bibitem[Edwards and Storkey(2016)]{edwards2015censoring}
Harrison Edwards and Amos Storkey.
\newblock Censoring representations with an adversary.
\newblock \emph{The International Conference on Learning Representations
  (ICLR)}, 2016.

\bibitem[Goodfellow et~al.(2014)Goodfellow, Pouget-Abadie, Mirza, Xu,
  Warde-Farley, Ozair, Courville, and Bengio]{goodfellow2014generative}
Ian Goodfellow, Jean Pouget-Abadie, Mehdi Mirza, Bing Xu, David Warde-Farley,
  Sherjil Ozair, Aaron Courville, and Yoshua Bengio.
\newblock Generative adversarial nets.
\newblock In \emph{Advances in neural information processing systems}, pages
  2672--2680, 2014.

\bibitem[Grgi{\'c}-Hla{\v{c}}a et~al.(2018)Grgi{\'c}-Hla{\v{c}}a, Zafar,
  Gummadi, and Weller]{grgic2018beyond}
Nina Grgi{\'c}-Hla{\v{c}}a, Muhammad~Bilal Zafar, Krishna~P Gummadi, and Adrian
  Weller.
\newblock Beyond distributive fairness in algorithmic decision making: Feature
  selection for procedurally fair learning.
\newblock In \emph{Thirty-Second AAAI Conference on Artificial Intelligence},
  2018.

\bibitem[Hardt et~al.(2016)Hardt, Price, Srebro, et~al.]{hardt2016equality}
Moritz Hardt, Eric Price, Nati Srebro, et~al.
\newblock Equality of opportunity in supervised learning.
\newblock In \emph{Advances in neural information processing systems}, page
  3315, 2016.

\bibitem[Huang et~al.(2007)Huang, Chen, and Wang]{huang2007credit}
Cheng-Lung Huang, Mu-Chen Chen, and Chieh-Jen Wang.
\newblock Credit scoring with a data mining approach based on support vector
  machines.
\newblock \emph{Expert systems with applications}, 33\penalty0 (4):\penalty0
  847--856, 2007.

\bibitem[Ibarrarán et~al.(2017)Ibarrarán, Medellín, Regalia, Stampini,
  Parodi, Tejerina, Cueva, and Vásquez]{ibarraran2017conditional}
Pablo Ibarrarán, Nadin Medellín, Ferdinando Regalia, Marco Stampini, Sandro
  Parodi, Luis Tejerina, Pedro Cueva, and Madiery Vásquez.
\newblock \emph{{How Conditional Cash Transfers Work}}.
\newblock Number 8159 in IDB Publications (Books). Inter-American Development
  Bank, 2017.
\newblock ISBN ARRAY(0x47b15000).
\newblock URL \url{https://ideas.repec.org/b/idb/idbbks/8159.html}.

\bibitem[Janocha and Czarnecki(2017)]{janocha2017loss}
Katarzyna Janocha and Wojciech~Marian Czarnecki.
\newblock On loss functions for deep neural networks in classification.
\newblock \emph{Conference on Theoretical Foundations of Machine Learning
  (TFML)}, 2017.

\bibitem[Joseph et~al.(2016)Joseph, Kearns, Morgenstern, and
  Roth]{joseph2016fairness}
Matthew Joseph, Michael Kearns, Jamie~H Morgenstern, and Aaron Roth.
\newblock Fairness in learning: Classic and contextual bandits.
\newblock In \emph{Advances in Neural Information Processing Systems}, pages
  325--333, 2016.

\bibitem[Kleinberg et~al.(2017)Kleinberg, Mullainathan, and
  Raghavan]{kleinberg2016inherent}
Jon Kleinberg, Sendhil Mullainathan, and Manish Raghavan.
\newblock Inherent trade-offs in the fair determination of risk scores.
\newblock \emph{Innovations in Theoretical Computer Science (ITCS)}, 2017.

\bibitem[Krishnapuram et~al.(2011)Krishnapuram, Yu, and
  Rao]{krishnapuram2011cost}
Balaji Krishnapuram, Shipeng Yu, and R~Bharat Rao.
\newblock \emph{Cost-sensitive Machine Learning}.
\newblock CRC Press, 2011.

\bibitem[Lichman et~al.(2013)]{lichman2013uci}
Moshe Lichman et~al.
\newblock Uci machine learning repository, 2013.

\bibitem[Logan(2014)]{logan2014diversity}
John Logan.
\newblock \emph{Diversity and disparities: America enters a new century}.
\newblock Russell Sage Foundation, 2014.

\bibitem[Ma et~al.(2019)Ma, Tschiatschek, Palla, Hernandez-Lobato, Nowozin, and
  Zhang]{ma2018eddi}
Chao Ma, Sebastian Tschiatschek, Konstantina Palla, Jose~Miguel
  Hernandez-Lobato, Sebastian Nowozin, and Cheng Zhang.
\newblock Eddi: Efficient dynamic discovery of high-value information with
  partial vae.
\newblock In \emph{International Conference on Machine Learning}, pages
  4234--4243, 2019.

\bibitem[Madras et~al.(2018)Madras, Creager, Pitassi, and
  Zemel]{madras2018learning}
David Madras, Elliot Creager, Toniann Pitassi, and Richard Zemel.
\newblock Learning adversarially fair and transferable representations.
\newblock In \emph{International Conference on Machine Learning}, pages
  3381--3390, 2018.

\bibitem[Mnih et~al.(2016)Mnih, Badia, Mirza, Graves, Lillicrap, Harley,
  Silver, and Kavukcuoglu]{mnih2016asynchronous}
Volodymyr Mnih, Adria~Puigdomenech Badia, Mehdi Mirza, Alex Graves, Timothy
  Lillicrap, Tim Harley, David Silver, and Koray Kavukcuoglu.
\newblock Asynchronous methods for deep reinforcement learning.
\newblock In \emph{International conference on machine learning}, pages
  1928--1937, 2016.

\bibitem[Noriega-Campero et~al.(2019)Noriega-Campero, Bakker, Garcia-Bulle, and
  Pentland]{noriega2018active}
Alejandro Noriega-Campero, Michiel Bakker, Bernardo Garcia-Bulle, and Alex
  Pentland.
\newblock Active fairness in algorithmic decision making.
\newblock \emph{Proceedings of AAAI / ACM Conference on Artificial
  Intelligence, Ethics, and Society}, 2019.

\bibitem[Obermeyer et~al.(2019)Obermeyer, Powers, Vogeli, and
  Mullainathan]{obermeyer447}
Ziad Obermeyer, Brian Powers, Christine Vogeli, and Sendhil Mullainathan.
\newblock Dissecting racial bias in an algorithm used to manage the health of
  populations.
\newblock \emph{Science}, 366\penalty0 (6464):\penalty0 447--453, 2019.
\newblock ISSN 0036-8075.
\newblock \doi{10.1126/science.aax2342}.
\newblock URL \url{https://science.sciencemag.org/content/366/6464/447}.

\bibitem[Phan et~al.(2019)Phan, Jin, Thai, Hu, and Dou]{phan2019preserving}
NhatHai Phan, Ruoming Jin, My~T Thai, Han Hu, and Dejing Dou.
\newblock Preserving differential privacy in adversarial learning with provable
  robustness.
\newblock \emph{arXiv preprint arXiv:1903.09822}, 2019.

\bibitem[Shim et~al.(2018)Shim, Hwang, and Yang]{shim2018joint}
Hajin Shim, Sung~Ju Hwang, and Eunho Yang.
\newblock Joint active feature acquisition and classification with
  variable-size set encoding.
\newblock In \emph{Advances in Neural Information Processing Systems}, pages
  1368--1378, 2018.

\bibitem[Verma and Rubin(2018)]{verma2018fairness}
Sahil Verma and Julia Rubin.
\newblock Fairness definitions explained.
\newblock In \emph{2018 IEEE/ACM International Workshop on Software Fairness
  (FairWare)}, pages 1--7. IEEE, 2018.

\bibitem[Vinyals et~al.(2015)Vinyals, Bengio, and Kudlur]{vinyals2015order}
Oriol Vinyals, Samy Bengio, and Manjunath Kudlur.
\newblock Order matters: Sequence to sequence for sets.
\newblock \emph{The International Conference on Learning Representations
  (ICLR)}, 2015.

\bibitem[Yang et~al.(2018)Yang, Brinton, Mittal, Chiang, and
  Lan]{yang2018learning}
Tsung-Yen Yang, Christopher Brinton, Prateek Mittal, Mung Chiang, and Andrew
  Lan.
\newblock Learning informative and private representations via generative
  adversarial networks.
\newblock In \emph{2018 IEEE International Conference on Big Data (Big Data)},
  pages 1534--1543. IEEE, 2018.

\bibitem[Zhang et~al.(2018)Zhang, Lemoine, and Mitchell]{zhang2018mitigating}
Brian~Hu Zhang, Blake Lemoine, and Margaret Mitchell.
\newblock Mitigating unwanted biases with adversarial learning.
\newblock In \emph{Proceedings of the 2018 AAAI/ACM Conference on AI, Ethics,
  and Society}, pages 335--340. ACM, 2018.

\end{thebibliography}
\bibliographystyle{plainnat}

\appendix

\setcounter{figure}{0}
\renewcommand{\thefigure}{SM\arabic{figure}}
\setcounter{table}{0}
\renewcommand{\thetable}{SM\arabic{table}}

\section{Set Encoder}\label{sec:setencoder}
A set encoder is used to encode arbitrary sets of features. The set encoder was introduced as part of the sequence-to-sequence framework in \citep{vinyals2015order}, while the authors in \citep{shim2018joint} adopt it for active feature-value acquisition. The set encoder has two parts: a \emph{reading block} and a \emph{processing block}. First, each feature is represented by a vector $\mathbf{u}_j=[x_j \; \mathcal{I}(j)]$ where $x_j$ is the feature-value and $\mathcal{I}(j)$ is a one-hot vector with $1$ at position $j$ and zeros elsewhere, allowing the network to incorporate coordinate information. The reading block embeds each vector $\mathbf{u}_j$ onto a memory vector $\mathbf{m}_j$ using a neural network with a shared set of parameters across all features $j\in \{1,\dots,d\}$. The processing block reads the memory (so all memory vectors) into an initial reading vector $\mathbf{r}_0=\frac{1}{N}\sum_j \mathbf{m}_j$ at processing step $0$. This vector $\mathbf{r}_0$ is padded with zeros and fed to an LSTM to compute an initial query vector $\mathbf{q}_0$. At each consecutive time step $t$ an attention weight for each memory vector $\mathbf{m}_i$ is computed using
\begin{align}
     a_{i,t}=\frac{\exp{(\mathbf{m}_i^{T}\mathbf{q}_t)}}{\sum_j \exp{(\mathbf{m}_j^{T}\mathbf{q}_t)}}
\end{align}
where $\mathbf{m}_i^{T}\mathbf{q}_t$ is the dot product of the memory and query vectors. Using the attention vector $\mathbf{a}_t$, we update the reading vector $\mathbf{r}_t=\sum_i a_{i,t} \mathbf{m}_i$ which we concatenate with the query vector and feed to the LSTM to compute the next query vector $\mathbf{q}_{t+1} =$LSTM$([\mathbf{q}_{t} \; \mathbf{r}_{t}])$. In turn, this new query vector is used to compute the new attention vector $\mathbf{a}_t$, We repeat this process for a fixed number of processing steps to achieve a final readout vector $\mathbf{r}_T$, which is subsequently fed to the classifiers and policy network. We refer to \citep{vinyals2015order} for a more detailed description over the encoder and experiments for different number of processing steps. Note that the attention mechanism guarantees that the final readout vector $\mathbf{r}_T$ retrieved from processing is invariant to different permutations of the features in the set.

\section{Architecture and training details}\label{sec:arch}
\paragraph{Architecture} We use two separate encoders $f_\theta$ and $f_\chi$ with the same architecture but different parameters where one feeds into the label classifier and one feeds into the adversary. The encoders consist of a memory block, a neural network with two hidden layers of 64-64 units that maps each feature value and its coordinate information to a 32-dimensional real-valued memory vector, and a processing block, an LSTM with 32 hidden units that performs 5 processing steps over the memory to obtain a final read vector. Both classifiers $g_\psi$ and $g_\omega$ and the policy network $\pi_\phi$ are realized as neural networks with two hidden layers of 64-64 units. The networks share the same architecture for both datasets and use rectified linear units (ReLUs) as activation functions. 

\paragraph{Pretraining}  In the first training phase, we train the encoders $f_\theta$ and $f_\chi$, and classifiers $g_\psi$ and $g_\omega$ with both the full set of features and randomly missing features. To obtain the partially missing feature sets, we drop each feature with probability $p\sim U(0,1)$, sampled once for instance to encourage different degrees of sparsity across instances. We train the models using the Adam optimizer with binary cross-entropy loss for 10,000 iterations and a batch size of 64. In each batch, half of the samples have randomly missing features and half contain the full feature set. We evaluate the AUC of the models on a validation set with partially missing features and save the models with the highest validation score for joint training.

\paragraph{Joint Training} In the second training phase, the policy network $\pi_\phi$ and the classifier $g_\psi$ are trained jointly for 10,000 iterations. We use n-step Q-Learning \citep{mnih2016asynchronous} with 4 steps and follow the implementation in \citep{shim2018joint} where multiple agents run in parallel and collect n-step experiences $(s_t, a_t, s_{t+1},a_{t+1},...,s_{t+n}, a_{t+n})$ using $\epsilon$-greedy exploration. We decrease $\epsilon$ linearly in the first 5,000 iterations from $1$ to $0.1$.  We train with 64 agents in parallel, one agent for one respective instance in a batch of 64. After collecting a running history of n-step experiences, $f_\theta$, $\pi_\phi$ and $g_\psi$ are jointly updated. The policy network $\pi_\phi$ and encoder $f_\theta$ are updated using gradient descent by backpropagating the squared loss $(Q(s_t, a_t)-R)^2$ of the estimated Q-values and the target Q-values. Q-values corresponding to actions of already acquired features are manually set to $-\infty$ to prevent the agent from selecting the same feature twice. To account for overestimation of Q-values and improve stability, we use a target Q network $\pi_{\phi'}$, which is a delayed copy of the online policy network $\pi_\phi$, that gets updated every 100 joint training iterations by copying the parameters of $\pi_\phi$. The estimated Q-values are defined as $Q(s_{t+n}, \arg\max_a{Q(s_{t+n}, a; \phi)}; {\phi'})$. The classifiers $g_\psi$ and $g_\omega$, together with encoders $f_\theta$ and $f_\chi$ are trained using the collected experiences of the agents. Each state $\{x_j\}_{j\in \mathcal{O}_t}$ in the history of experiences represents a partial feature vector that is used in combination with the true label to update the classifier. In line with the pretraining phase, the classifier is trained using binary cross-entropy loss and the Adam optimizer. 

\end{document}